\documentclass[runningheads]{llncs}
\usepackage{hyperref} 
\usepackage{esvect}
\usepackage{pdfpages}
\usepackage[T1]{fontenc}
\usepackage[T1]{fontenc}
\usepackage{graphicx}

\usepackage{xcolor}
\usepackage{float}
\usepackage{url}
\usepackage{xcolor}
\usepackage{longtable}
\usepackage{array}
\usepackage{multirow}
\usepackage{placeins}
\usepackage{graphicx}
\usepackage{booktabs}
\usepackage{amsmath}
\usepackage{amssymb}
\usepackage{xcolor}
\usepackage{hyperref}
\usepackage{algorithm}
\usepackage{algpseudocode}
\usepackage{cite}
\usepackage[dvipsnames]{xcolor}

\definecolor{revblue}{RGB}{0,70,200}

\usepackage{amsmath}

\makeatletter
\renewcommand\subparagraph{\@startsection{subparagraph}{5}{\z@}%
  {-6pt plus -1pt minus -1pt}{-0.5em}{\normalfont\itshape}}
\makeatother
\usepackage{titlesec}

\titlespacing*{\section}{0pt}{8pt plus 2pt minus 2pt}{4pt plus 1pt minus 1pt}
\titlespacing*{\subsection}{0pt}{6pt plus 2pt minus 2pt}{3pt plus 1pt minus 1pt}
\titlespacing*{\subsubsection}{0pt}{4pt plus 1pt minus 1pt}{2pt plus 1pt minus 1pt}
\hypersetup{
    colorlinks=true,
    linkcolor=blue,
    citecolor=blue,
    urlcolor=blue
}

\makeatletter
\renewcommand{\and}{\unskip\ and\ \ignorespaces}
\makeatother

\begin{document}
\title{Reading Beyond OCR Noise: Robust Cross-Lingual RAG for Legislative Archives}
\titlerunning{Robust Cross-Lingual RAG for Legislative Archives}
%
\author{
Radha Krishna Deshpande
\and
Ajoy Mondal
\and
Akash Manna
\and
C V Jawahar}

\authorrunning{R K Deshpande et al.}
%
\institute{
CVIT,  International Institute of Information Technology, Hyderabad, India
\email{(radhakrishna.d,akash.manna)@research.iiit.ac.in , (ajoy.mondal,jawahar)@iiit.ac.in}
}
\maketitle              
\begin{abstract}
India's historical public archives present severe computer vision and document image analysis challenges: decades of parliamentary records survive only as degraded, multi-column scanned pages featuring mixed scripts, complex layouts, and dense visual noise where OCR failure is the primary obstacle to information retrieval. We present an end-to-end framework that couples vision-based layout segmentation and multi-script OCR with an agentic retrieval-augmented generation (RAG) pipeline. Our approach specifically addresses how visual degradation and layout fragmentation propagate into text extraction errors, using an iterative agentic loop to navigate around OCR noise and layout ambiguities before generating answers. Evaluated on the 200,000-page Punjab Vidhan Sabha archive, our system demonstrates robust handling of historical document degradation, outperforming standard text-only baselines. The code is publicly available at \href{https://github.com/radhakrishnadeshpande/Reading-Beyond-OCR-Noise-Robust-Cross-Lingual-RAG-for-Legislative-Archives}{GitHub Repo}.

\end{abstract}
\section{Introduction}

Historical legislative archives are difficult to process because the main challenge is often the document image itself. Faded text, irregular layouts, marginal notes, and mixed scripts can introduce errors during layout analysis and OCR, which then propagate to retrieval and question answering \cite{zhang2024ohrbench,piryani2025multiocr,romein2025transforming}. This is particularly challenging for large archives containing Hindi, Punjabi, and English documents with varying print quality and complex layouts.

Rather than treating OCR as a separate preprocessing step, we study the complete pipeline from document images to retrieval and answer generation. Our framework combines layout-aware OCR with an agentic retrieval loop that can check the retrieved evidence, reformulate the query, and search again when the available context is insufficient. This helps keep generated answers grounded in the original document evidence.

Retrieval-augmented generation (RAG) was introduced to combine language models with external knowledge sources \cite{lewis2020rag} and has since been extended to iterative and agent-based retrieval \cite{gao2024survey,singh2025agentic,mishra2025sok}. Recent document retrieval methods have also moved toward vision-based representations, reducing reliance on OCR for document retrieval \cite{faysse2024colpali,ma2024dse}. Multi-page document understanding further shows the importance of preserving visual information during retrieval and reasoning \cite{cho2024m3docrag}. At the same time, OCR benchmarks have shown that multimodal language models still face difficulties with text recognition, particularly for multilingual and visually challenging documents \cite{liu2024ocrbench}, while recent work demonstrates that OCR errors can significantly affect downstream question answering \cite{piryani2025multiocr}. Language-model-based OCR correction has therefore become an active research direction, especially for historical documents \cite{bourne2025clocrccontextleveragingocr,thomas2024leveraging}.

These developments motivate an end-to-end approach for historical legislative archives in which visual processing, retrieval, and answer generation are evaluated together. We focus on the Punjab Vidhan Sabha proceedings, a large multilingual archive spanning 1950--2024. Our system uses open-weight models throughout the pipeline so that the complete workflow can be deployed locally and the retrieved evidence can be traced back to the source pages.

In this paper, we contribute:
\begin{itemize}
\item an agentic RAG framework that combines layout-aware OCR, hybrid retrieval, and iterative evidence checking for a multilingual legislative archive;
\item an evaluation of cross-lingual question answering on a large-scale historical document collection;
\item an analysis of how OCR and layout errors affect retrieval and answer quality, together with an ablation of the agentic retrieval loop.
\end{itemize}

\section{Related Work}

Work on multilingual archive navigation sits at the intersection of several distinct lines of research: retrieval-augmented generation, agentic search, multilingual text embeddings, and multimodal document retrieval. None of these paradigms were originally optimized for a corpus precisely like ours, making it necessary to examine their contributions and limitations regarding noisy, multi-script legislative archives.

\subsection{Agentic and Multilingual Retrieval-Augmented Generation}

Retrieval-augmented generation originated from the premise that language models perform better when granted access to external knowledge rather than depending solely on parametric memory \cite{lewis2020rag}. Gao et al.\ categorize RAG frameworks into naive, advanced, and modular designs, emphasizing that iterative query reformulation yields greater accuracy gains than merely scaling retriever capacity \cite{gao2024survey}. 

Agentic RAG advances this by treating retrieval as an adaptive, multi-step agent loop rather than a static query-response pipeline \cite{singh2025agentic,mishra2025sok}. Architectures that dynamically navigate between keyword indexing, dense semantic search, and document-level browsing offer crucial flexibility for decades-old legislative records \cite{ayanami2025arag}. While agentic orchestration has been successfully demonstrated on historical collections like the CIA's declassified archives, those implementations typically remained monolingual without addressing script-switching constraints \cite{cerny2026multistage}.

Cross-lingual and multilingual RAG present distinct structural hurdles. Chirkova et al.\ demonstrated that maintaining output consistency requires robust prompt design rather than auxiliary translation steps \cite{chirkova2024multilingual}, while Liu et al.\ observed performance degradation in cross-lingual retrieval tasks \cite{liu2025xrag}. Conversely, Li et al.\ highlighted that incorporating multilingual evidence can stabilize reasoning in ambiguous contexts \cite{li2024bordirlines}. Influenced by these findings, our system indexes Hindi, Punjabi, and English sources concurrently. Foundation models such as Multilingual-E5-Large provide strong cross-lingual mappings, serving as an important precedent for our multi-script workflow \cite{wang2024multilingual,sener2024advanced}.

\subsection{OCR and Historical Multilingual Document Processing}

Historical document retrieval requires careful accounting for OCR degradation. Zhang et al.\ quantified how noise introduced during text extraction degrades downstream RAG performance \cite{zhang2024ohrbench}, and Piryani et al.\ showed that these errors compound significantly in non-Latin and multi-script environments like Punjabi Gurmukhi and Hindi Devanagari \cite{piryani2025multiocr}. 

Recent post-correction methodologies utilize LLMs to clean historical text and generalize better than traditional sequence models \cite{thomas2024leveraging}. Bourne demonstrated that injecting domain-specific historical context further improves entity recognition for specialized legislative terminology \cite{bourne2025clocrccontextleveragingocr}. Romein et al.\ confirm that structurally complex, multi-script documents remain a primary challenge in AI-assisted archive digitization \cite{romein2025transforming}.

\subsection{Multimodal and Vision-Language Document Retrieval}

Traditional text-only pipelines often discard critical page layout structures such as tables, headers, and column boundaries, motivating a shift toward multimodal document retrieval. Ma et al.\ review this paradigm shift toward vision-language document encodings \cite{ma2025scaling}. Frameworks such as ColPali \cite{faysse2024colpali}, DSE \cite{ma2024dse}, and advanced visual embedding architectures like NetraEmbed \cite{kolavi2025m3dr} encode document images directly into dense vector spaces, preserving layout semantics and layout structures across multiple languages. Cho et al.\ extended this to multi-page retrieval frameworks \cite{cho2024m3docrag}, while Liu et al.\ emphasize the integration of sparse lexical search with dense visual embeddings as an optimal trajectory for complex document repositories \cite{liu2025mllms}. Our optimized pipeline leverages these insights by pairing robust multilingual textual dense representations with an adaptive agentic loop tailored for historical legislative document search.
\section{Methodology}
\subsection{Query Management and Multi-Lingual Processing}
\begin{figure}[H]
    \centering
    \includegraphics[width=0.8\linewidth]{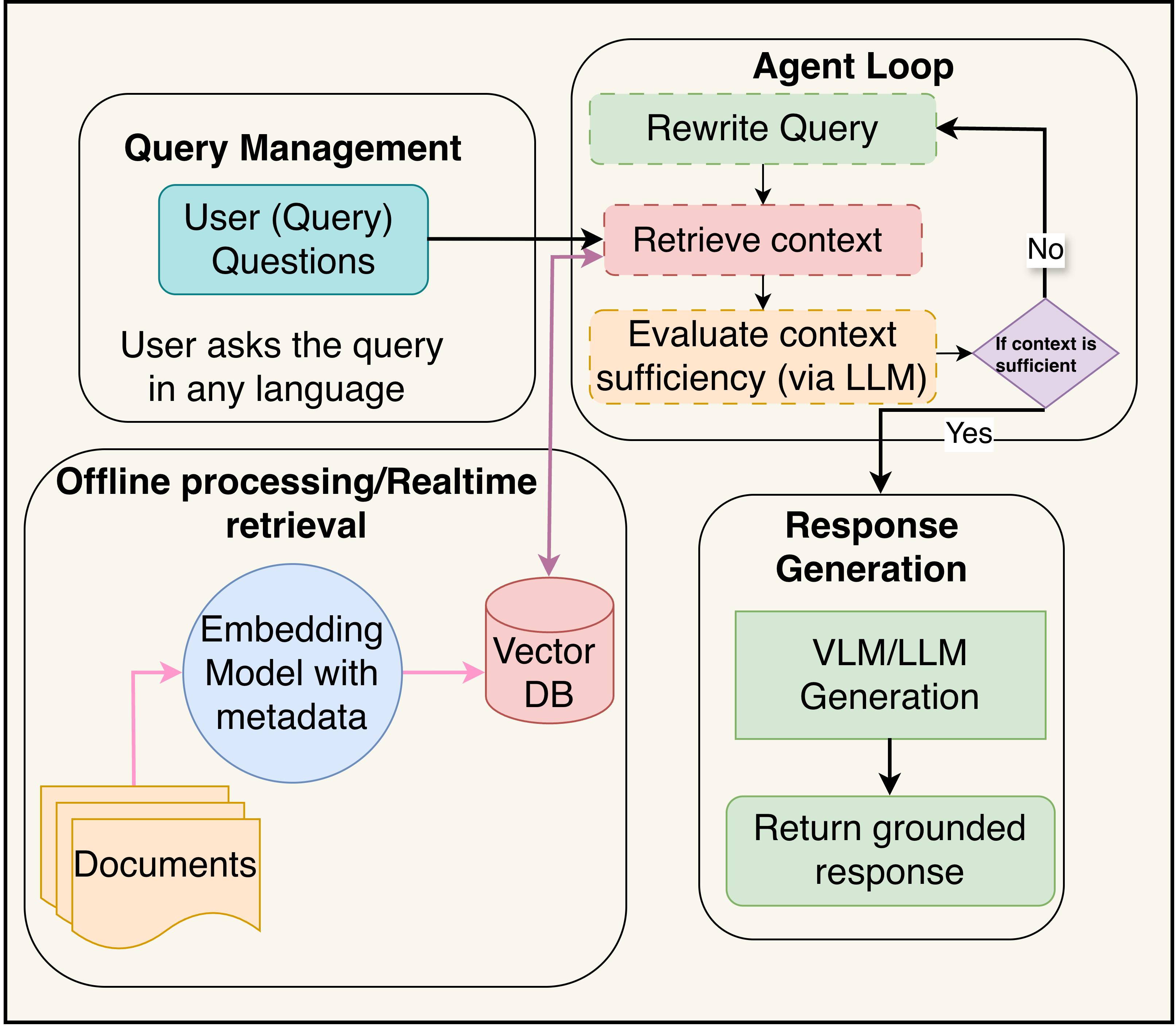}
    \caption{Architecture of the proposed Agentic Retrieval-Augmented Generation (RAG) system. The framework features multi-lingual query processing, an iterative LLM-driven evaluation loop to ensure context sufficiency, and a grounded response generation pipeline.}
    \label{fig:system_architecture}
\end{figure}

The system initiates by receiving a user query, denoted as $q$, in its native script and language. Rather than relying on intermediate machine translation, our architecture processes the query natively against a multilingual semantic vector space, preserving domain-specific legislative terminology and regional context across Hindi, Punjabi, and English. Fig.~\ref{fig:system_architecture} shows the pipeline: the query is embedded natively, the agent loop checks context sufficiency and rewrites the query if needed, and the generator produces the grounded answer.

\subsection{Offline Processing and Real-Time Retrieval}
The foundation of the retrieval system relies on a corpus of documents, $\mathcal{D} = \{d_1, d_2, \dots, d_n\}$. During the offline phase, an embedding model maps these documents into a dense vector space alongside their associated metadata. Let $E(\cdot)$ represent the multilingual embedding function. The resulting document embeddings are stored systematically within a Vector Database.

For real-time retrieval, the native query $q$ is embedded directly to produce a query vector $v_q = E(q)$. The system queries the Vector Database to retrieve the top-$k$ most relevant context chunks, $C = \{c_1, c_2, \dots, c_k\}$, by calculating the similarity score between $v_q$ and the document vectors $v_{d_i}$. 
\begin{equation} \operatorname{sim}(q,d_i) = \frac{v_q^\top v_{d_i}} {\|v_q\|_2 \|v_{d_i}\|_2}. \end{equation}

\subsection{Iterative Agent Loop for Context Evaluation}
A defining feature of this architecture is the autonomous Agent Loop, designed to mitigate hallucinations and ensure the retrieved context is adequately informative despite historical OCR degradation. Once the initial context $C$ is retrieved, a dedicated evaluation LLM analyzes its sufficiency to answer $q$ based on an information gain threshold function:

\begin{equation}
\mathcal{G}(C, q) = \sum_{j=1}^{|q|} \mathbb{I}(w_j \in \text{Tokens}(C))
\label{eq:information_gain}
\end{equation}

\begin{itemize}
    \item If the context score satisfies the sufficiency threshold, the workflow proceeds directly to the generation phase.
    \item If the context is deemed insufficient, the agent triggers an adaptive query rewriting mechanism to generate an optimized query variant $q'$. The optimization objective minimizes semantic ambiguity:
    \begin{equation}
    q' = \arg\max_{\tilde{q}} \text{Sim}(E(\tilde{q}), E(q)) - \lambda \mathcal{L}_{\text{OCR\_noise}}(\tilde{q})
    \label{eq:query_rewrite}
    \end{equation}
    
    Here, $\lambda$ balances semantic similarity to $q$ against $\mathcal{L}_{\text{OCR\_noise}}(\tilde{q})$, a conceptual penalty on query terms likely to be missed due to OCR errors in the indexed pages, which in practice is approximated by prompting the rewriting LLM to preserve the intent of $q$ while rephrasing terms that failed to retrieve evidence.
    The system then loops back to the retrieval step using this refined query to fetch a more robust contextual foundation.
\end{itemize}

\subsection{Grounded Response Generation}
Upon validating context sufficiency, the generation LLM synthesizes a grounded response directly in the user's native language. This generation is strictly conditioned on both the native query and the validated context, maximizing factual grounding without translation intermediaries. The probability of generating the response sequence of tokens $w$ can be modeled as:

\begin{equation}
P(R \mid q, C) = \prod_{t=1}^{T} P(w_t \mid w_{<t}, q, C)
\label{eq:response_gen}
\end{equation}

\subsection{Algorithm: Native-Language Agentic RAG Loop}
\label{sec:agentic_rag_algo}

\begin{algorithm}[H]
\caption{Native-Language Multi-Step Agentic Retrieval}
\label{alg:agentic_rag}
\begin{algorithmic}[1]
\Require User query $q$, Vector DB collection $\mathcal{V}$, Maximum loops $L$
\Ensure Grounded response $R$ in native language
\State $current\_query \gets q$
\For{$step = 1$ to $L$}
    \State $C \gets \mathcal{V}.\text{query}(current\_query, k=5)$
    \State $is\_sufficient, eval\_log \gets \text{LLM\_Evaluate}(current\_query, C[0])$
    \If{$is\_sufficient$ or $step == L$}
        \State \textbf{break}
    \EndIf
    \State $current\_query \gets \text{LLM\_Rewrite}(current\_query, eval\_log)$
\EndFor
\State $R \gets \text{LLM\_Generate}(q, C)$
\State \textbf{return} $R$
\end{algorithmic}
\end{algorithm}

\section{Experiments and Results}

\subsection{Dataset}
The dataset used in this research consists of Punjab Vidhan Sabha (legislative assembly of Punjab state) documents collected from archival records spanning from the 1950s to 2024. The collection includes assembly proceedings, debates, reports, notices, and other official legislative documents published across different time periods.

The records were digitized from bound volumes held by [\url{https://pvs.neva.gov.in/}], a process that was carried out incrementally over several decades rather than in a single digitization effort. This matters for the corpus's characteristics: pages from the 1950s and 1960s were typically typeset on older printing equipment and later scanned at lower resolution, often with visible ink bleed-through, faded typeface, and skewed page boundaries, while more recent decades (post-2020) are comparatively cleaner but still exhibit dense multi-column layouts and inconsistent formatting across sessions.

\begin{figure}
    \centering
    \includegraphics[width=1.0\linewidth]{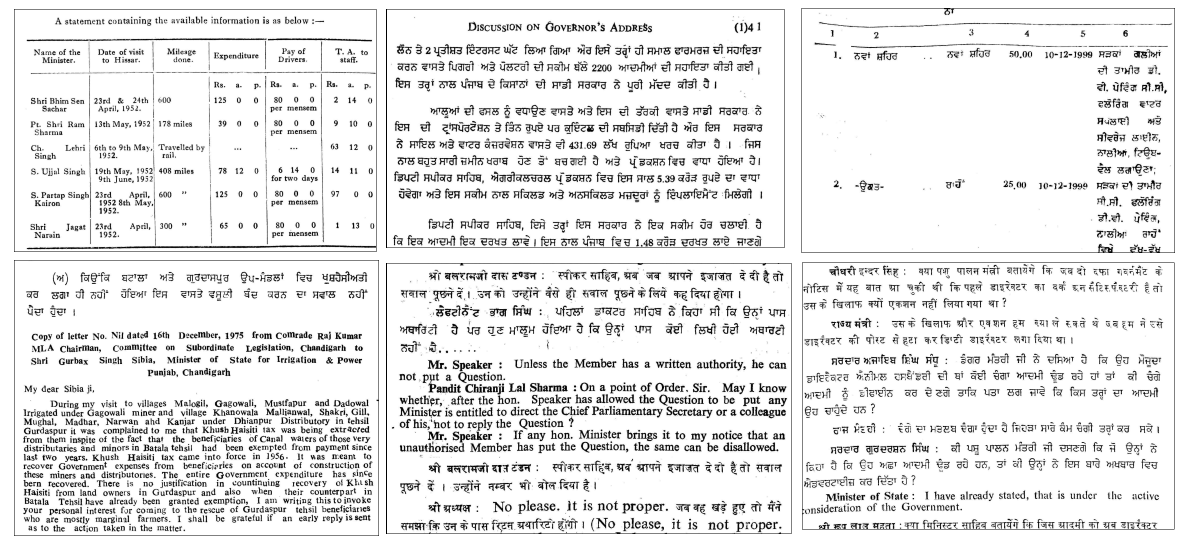}
    \caption{Overview of dataset pages featuring tabular data, multi-scripts, and faded text characteristics.}
    \label{fig:dataset}
\end{figure}

Documents in the corpus are predominantly in Hindi (Devanagari) and Punjabi (Gurmukhi), with English appearing throughout, particularly in earlier-era records and administrative headers; it is common for a single page to mix two or all three languages, for instance an English-language question header followed by a Punjabi response, or bilingual tabular data. The collection totals approximately 200,000 pages a covering assembly sessions, starred and unstarred questions, budget and financial statements, committee reports, and procedural notices. Fig.~\ref{fig:dataset} shows representative pages illustrating the multi-column layouts, mixed scripts, and visual degradation typical of the corpus.

Given the scale of the archive, we did not attempt to hand-annotate the full collection. Instead, we constructed a stratified evaluation benchmark of 1,000 pages and 200 gold-annotated question-answer pairs, sampled to balance representation across decade, script, and document type, so that evaluation results reflect the corpus's actual heterogeneity rather than only its cleaner, more recent portion. To isolate the contribution of the vision pipeline specifically, we additionally hand-transcribed a smaller subset of 20 pages, selected to span a range of degradation levels, as ground truth for OCR quality evaluation.

\subsection{Experimental Setup}

We ran our experiments on a single Nvidia RTX-4060 (16 GB VRAM). Our evaluation framework implemented a robust local storage backend using ChromaDB alongside an adaptive agentic retrieval architecture driven by \texttt{BAAI/bge-m3} for dense and hybrid token-level multilingual embeddings (supporting English, Punjabi Gurmukhi, and Hindi Devanagari).

As described in Section~\ref{sec:agentic_rag_algo}, the retrieval index spans the full 200,000-page Punjab Vidhan Sabha archive. All quantitative results in this section are computed against our stratified evaluation benchmark of 1,000 pages and 200 gold-annotated queries, drawn from this corpus and balanced across decade, script, and document type. Additionally, to isolate vision-side performance, we hand-transcribed a subset of 20 pages and evaluated multiple layout segmentation and OCR model combinations against this ground truth. 

\subsection{Metrics}
\label{subsec:metrics}

We evaluate the proposed system at three levels: OCR recognition,
document retrieval, and answer quality.

\textbf{OCR Recognition:}
We use Character Recognition Rate (CRR) and Word Recognition Rate (WRR)
to evaluate OCR performance. Both measures are derived from the
Levenshtein edit distance, which counts substitutions, deletions, and
insertions required to transform the predicted sequence into the
reference sequence \cite{icdar2024rdtag,ihtr2023}.

The Character Recognition Rate is defined as
\begin{align}
\mathrm{CRR}
&= 1-\frac{S_c+D_c+I_c}{N_c} \\
&= \frac{N_c-S_c-D_c-I_c}{N_c},
\end{align}
where $S_c$, $D_c$, and $I_c$ denote the number of character
substitutions, character deletions, and character insertions,
respectively, and $N_c$ denotes the number of characters in the
reference text.

Similarly, the Word Recognition Rate is defined as
\begin{align}
\mathrm{WRR}
&= 1-\frac{S_w+D_w+I_w}{N_w} \\
&= \frac{N_w-S_w-D_w-I_w}{N_w},
\end{align}
where $S_w$, $D_w$, and $I_w$ denote the number of word substitutions,
word deletions, and word insertions, respectively, and $N_w$ denotes
the number of words in the reference text \cite{icdar2024rdtag,ihtr2023}.

\textbf{Retrieval:}
We evaluate retrieval effectiveness using Mean Reciprocal Rank (MRR)
and Hit Rate at $k$ (Hit@k). MRR measures the reciprocal rank of the
highest-ranked relevant document for each query and averages this value
over all queries \cite{voorhees1999trec}:
\begin{align}
\mathrm{MRR@1}
&= \frac{1}{|Q|}
\sum_{i=1}^{|Q|}
\frac{1}{\mathrm{rank}_i}.
\end{align}

Hit@k measures the proportion of queries for which at least one relevant
document is retrieved within the top-$k$ ranked results:
\begin{align}
\mathrm{Hit@k}
&= \frac{1}{|Q|}
\sum_{i=1}^{|Q|}
\mathbb{I}(\mathrm{rank}_i \leq k).
\end{align}

Here, $Q$ denotes the set of evaluation queries, $|Q|$ denotes the
number of queries, $\mathrm{rank}_i$ denotes the rank of the first
relevant document for query $i$, and $\mathbb{I}(\cdot)$ is the indicator
function that returns $1$ when the condition is satisfied and $0$
otherwise.

\textbf{Answer Quality:}
We use the F1 score, which combines Precision and Recall through their
harmonic mean:
\begin{align}
F1 = \frac{2PR}{P+R},
\end{align}
where $P$ denotes Precision and $R$ denotes Recall. Precision measures
the proportion of predicted answer tokens that are correct, while Recall
measures the proportion of reference answer tokens that are correctly
identified. The F1 score therefore provides a balanced measure of
answer quality \cite{chinchor1992muc}.

\subsection{Retrieval Performance Comparison}

Our primary goal was to evaluate and compare the retrieval performance of various multilingual embedding models and our proposed optimization framework on the domain-specific Punjab Vidhan Sabha (PVS) benchmark.

As shown in Table~\ref{tab:pvs_performance}, token-level and smaller cross-lingual models like MuRIL and LaBSE struggle heavily with the complex, multi-script historical documents, yielding near-zero or low MRR@1 scores. Larger dense retrievers such as Multilingual-E5-Large and NetraEmbed scale better, achieving Hit@3 scores above 0.81. However, BGE-M3 and our optimized pipeline establish the strongest baseline and performance standard, successfully navigating historical OCR degradation and mixed-script text to secure top-tier retrieval accuracy.

\begin{figure}[H]
    \centering
    \includegraphics[width=1.0\textwidth]{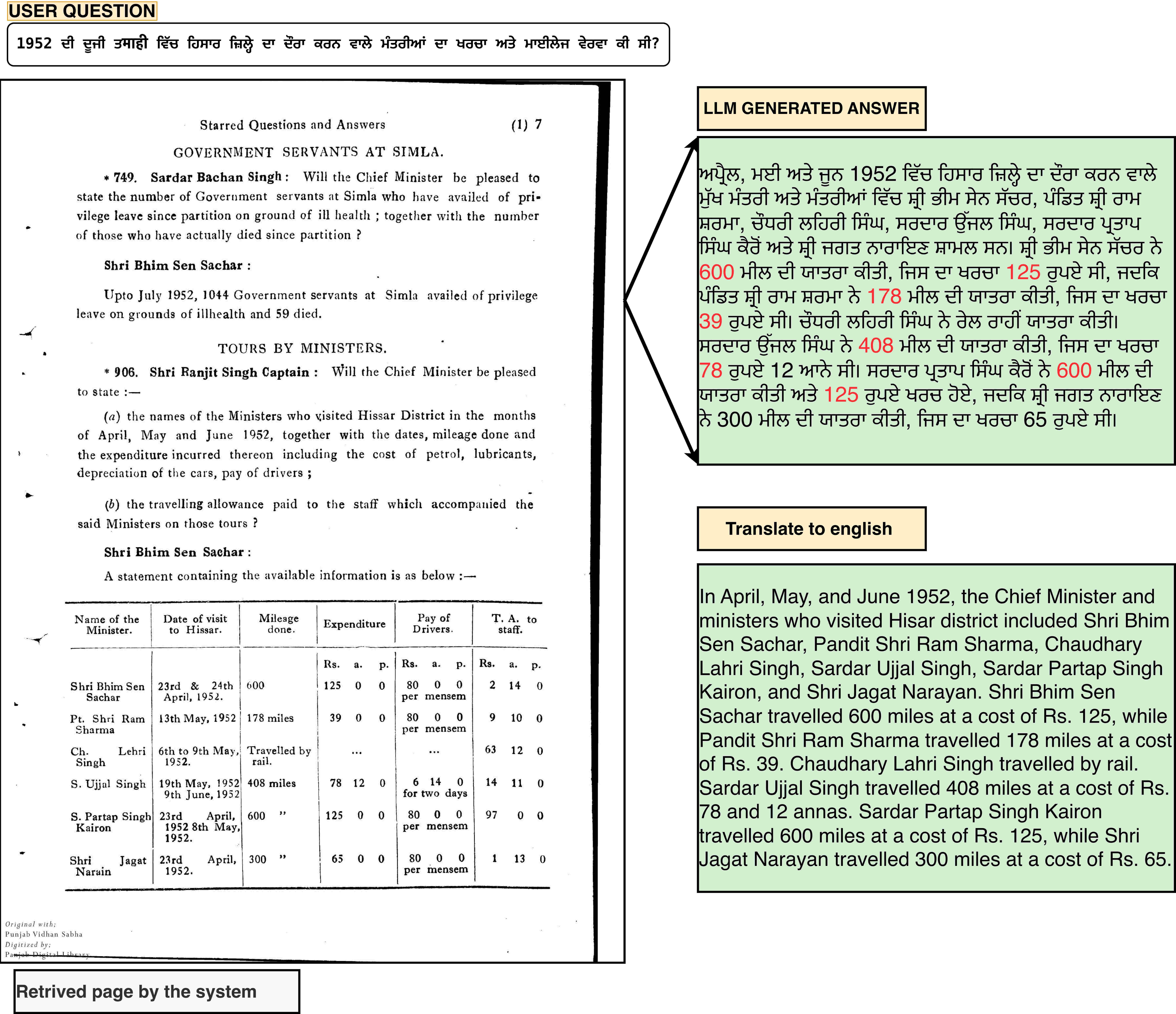}
    \caption{End-to-end cross-lingual RAG pipeline for  legislative documents, from the user query and document retrieval to grounded answer generation in Punjabi and translated to english for the reference.}
    \label{fig:cross_lingual}
\end{figure}
\begin{figure}[H]
    \centering
    \includegraphics[width=1.0\linewidth]{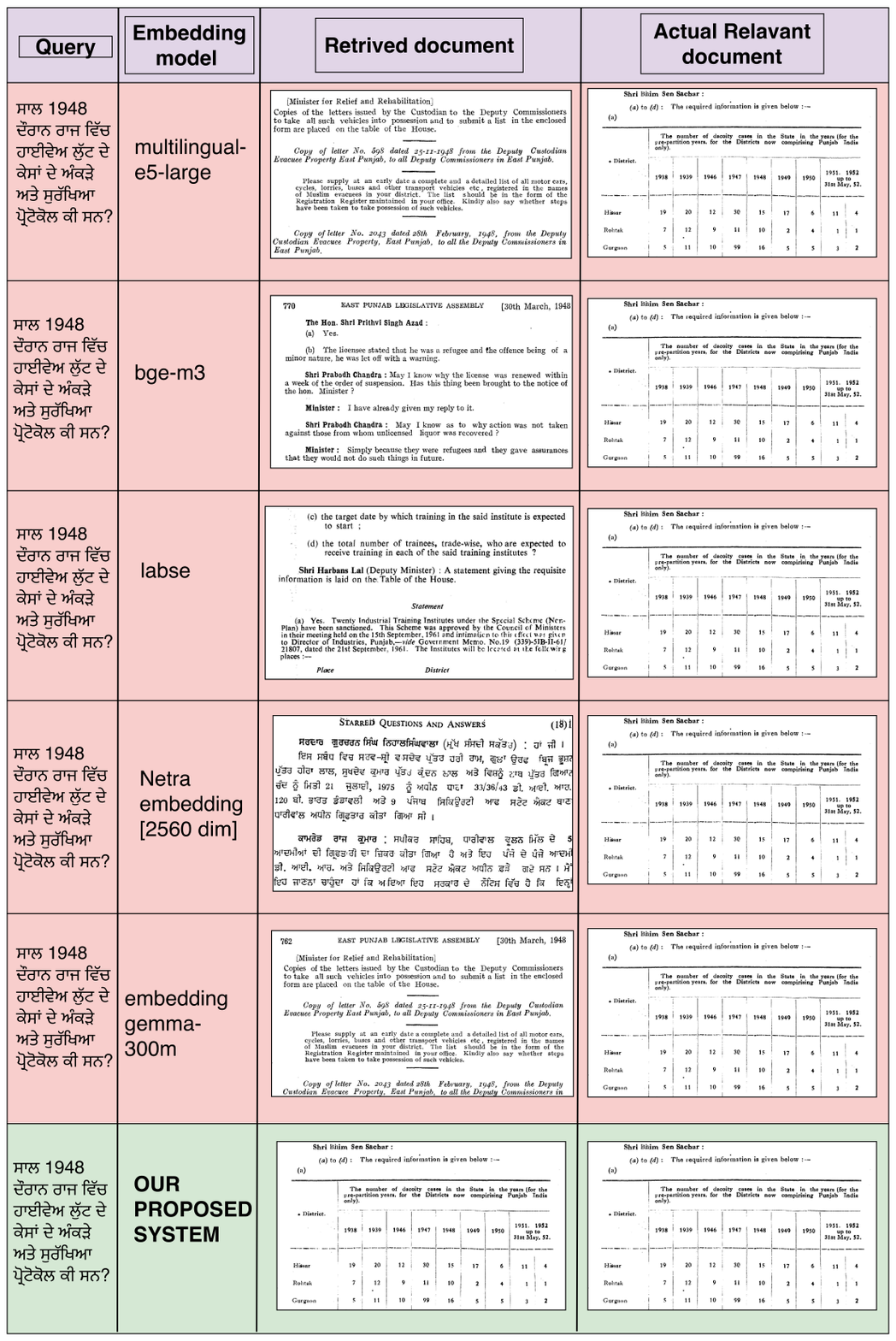}
    \caption{Qualitative comparison of cross-lingual embedding models for historical legislative document retrieval. The same Punjabi query is given to multilingual-e5-large, BGE-M3, LaBSE, Netra embedding (2560 dimensions), and EmbeddingGemma-300M. The retrieved document from each model is compared with the ground-truth relevant document shown in the rightmost column. The retrieved results demonstrate differences in cross-lingual semantic matching, with several models returning visually and semantically unrelated legislative records. In contrast, our proposed system retrieves the relevant document containing the required information, demonstrating its effectiveness for retrieval from multilingual historical archives.}
    \label{fig:embeddingmodel}
\end{figure}

Fig.~\ref{fig:cross_lingual} shows the full pipeline from query to grounded answer , and Fig.~\ref{fig:embeddingmodel} compares the pages retrieved by each model for one Punjabi query.

\begin{table}[htbp]
\centering
\caption{PVS Retrieval Evaluation on the 1,000-page / 200-query stratified benchmark (indexed against the full 200,000-page archive).}
\label{tab:pvs_performance}
\resizebox{\textwidth}{!}{%
\begin{tabular}{l c c c c c}
\toprule
\textbf{Model \& Pipeline Setup} & \textbf{Eval.\ Benchmark} & \textbf{MRR@1} & \textbf{Hit@3} & \textbf{Hit@5} & \textbf{Prec@3} \\
\midrule
MuRIL & 1,000 Pages / 200 Queries & 0.000 & 0.005 & 0.015 & 0.002 \\
LaBSE & 1,000 Pages / 200 Queries & 0.320 & 0.510 & 0.590 & 0.170 \\
Embedding-gemma (300M) & 1,000 Pages / 200 Queries & 0.525 & 0.735 & 0.775 & 0.245 \\
NetraEmbed (dim=768) & 1,000 Pages / 200 Queries & 0.580 & 0.816 & 0.890 & 0.272 \\
NetraEmbed (dim=1536) & 1,000 Pages / 200 Queries & 0.598 & 0.835 & 0.905 & 0.278 \\
NetraEmbed (dim=2560) & 1,000 Pages / 200 Queries & 0.605 & 0.843 & 0.912 & 0.281 \\
Multilingual-E5-Large & 1,000 Pages / 200 Queries & 0.620 & 0.855 & 0.915 & 0.285 \\
Standard BGE-M3 & 1,000 Pages / 200 Queries & 0.775 & 0.900 & 0.940 & 0.300 \\
\textbf{Ours (Optimized Pipeline)} & \textbf{1,000 Pages / 200 Queries} & \textbf{0.810} & \textbf{0.925} & \textbf{0.955} & \textbf{0.325} \\
\bottomrule
\end{tabular}%
}
\end{table}

\section{Ablation}

\subsection{OCR and Layout Pipeline Evaluation}
\label{sec:ocr_analysis}
Our production pipeline combines a \textbf{fine-tuned YOLO}
\cite{jocher2023yolo} layout model with our CRNN-based OCR recognizer.
We compare it with a \textbf{CRAFT}-based layout model
\cite{baek2019craft} using the same CRNN recognizer, along with Google
OCR and Surya OCR.

\begin{table}
\centering
\caption{OCR pipeline comparison on the 20-page ground-truth subset.}
\label{tab:ocr_pipeline}
\begin{tabular}{l l c c}
\toprule
\textbf{Layout Model} & \textbf{OCR Model} & \textbf{CRR (\%)} & \textbf{WRR (\%)} \\
\midrule

Gemini-3.1  & Gemini-3.1 & 99.7 & 89.6 \\
Surya2 & Surya2 & 99.6 & 88.9 \\
Sarvam & Sarvam & 99.5 & 89.4 \\
\textbf{YOLO (Fine-tuned)} & \textbf{CRNN (Ours)} & \textbf{95.67} & \textbf{80.03} \\
CRAFT & CRNN (Ours) & 89.00 & 60.35 \\
\bottomrule
\end{tabular}
\end{table}

The fine-tuned YOLO + CRNN pipeline achieves 95.67\% CRR and 80.03\% WRR,
compared with 89.00\% CRR and 60.35\% WRR for CRAFT + CRNN. Since the
same CRNN recognizer is used in both pipelines, the improvement shows
the effect of the layout model on the quality of the extracted text.
Google OCR and Surya OCR achieve higher recognition scores, while our
pipeline remains fully local and provides greater control for processing
the large archive.
\begin{table}
\centering
\caption{PVS Retrieval Evaluation using different OCR models on the 200-query benchmark.}
\label{tab:pvs_ocr_comparison}
\begin{tabular}{l l c c c c}
\toprule
\textbf{OCR Model} & \textbf{MRR@1} & \textbf{Hit@3} & \textbf{Hit@5} & \textbf{Prec@3} \\
\midrule
Gemini-3.1  & 0.825 & 0.920 & 0.960 & 0.3533 \\
Sarvam & 0.820 & 0.935 & 0.955 & 0.325 \\
Surya2  & 0.815 & 0.925 & 0.950 & 0.305 \\
\textbf{Ourocr} & \textbf{0.810} & \textbf{0.925} & \textbf{0.955} & \textbf{0.325} \\
\bottomrule
\end{tabular}
\end{table}

\subsection{Ablation on Agent Loop Iterations}
\label{sec:ablation_agent_loop}

To quantify the contribution of the iterative agentic loop, we vary the maximum number of retrieval--rewrite iterations using a fixed 200-query evaluation dataset, $L \in \{1, 2, 3\}$, and measure retrieval quality (MRR@1, Precision@3) alongside end-to-end answer correctness (Answer F1-Score) on the same 200-query benchmark. $L=1$ corresponds to a single-pass, non-agentic retrieval baseline with no query rewriting.Fig.~\ref{fig:ablation_iterations} illustrates how iterative query rewriting corrects retrieval errors. In the first two iterations, the system retrieves irrelevant documents, but after further query refinement, the third iteration successfully retrieves the correct document containing the required information.

\begin{figure}[H]
    \centering
    \begin{minipage}{0.5\textwidth}
        \centering
        \includegraphics[width=\linewidth]{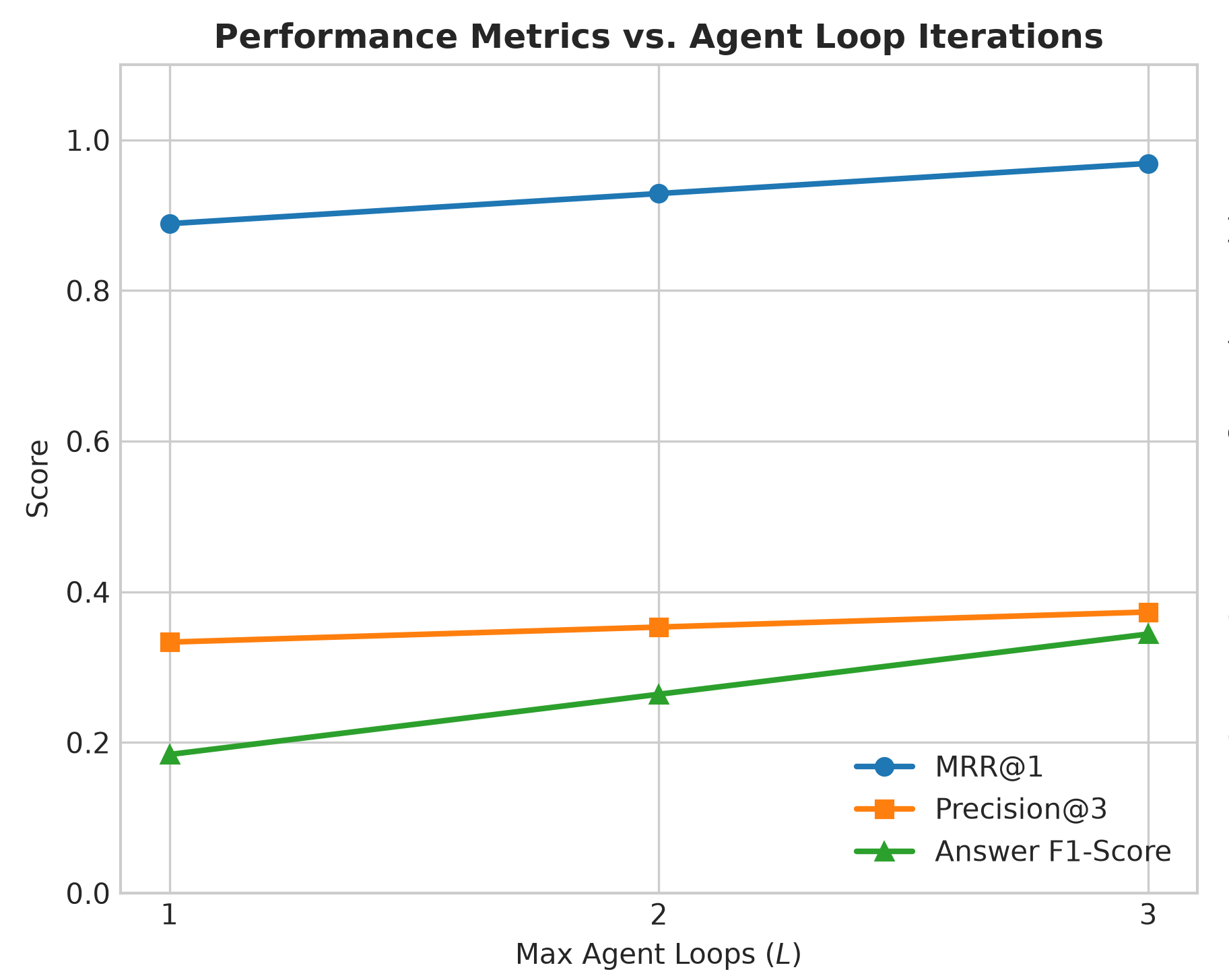}
        \caption{Performance metrics vs Iteration loops of the proposed system across different agent loop iterations. The results show how the system performance changes as the number of iterations increases.}
        \label{fig:performance_metrics}
    \end{minipage}
    \hfill
    \begin{minipage}{0.45\textwidth}
        \centering
        \includegraphics[width=\linewidth]{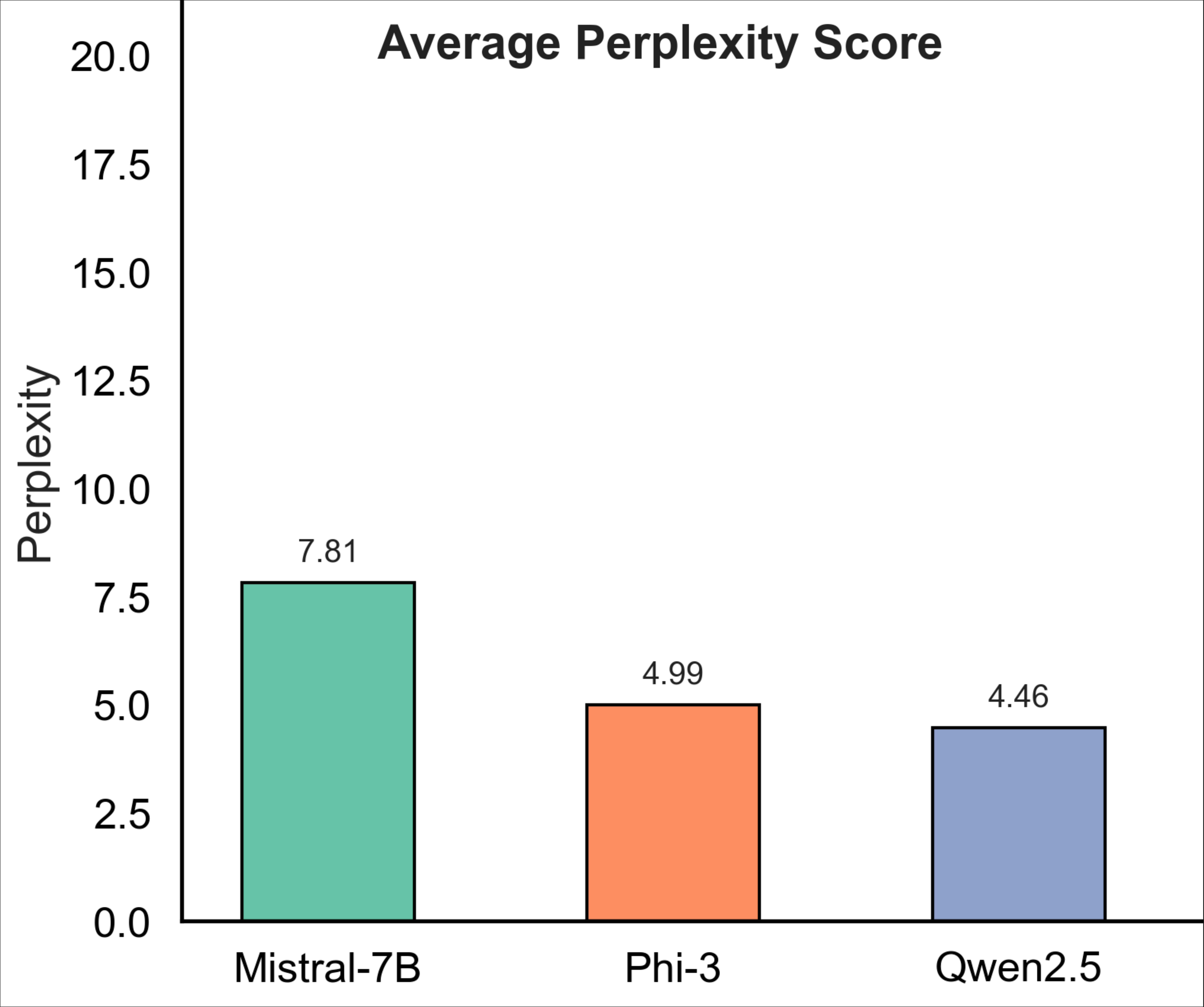}
        \caption{Average Perplexity scores of the language models on multi-script documents. Qwen2.5 achieves the lowest perplexity (4.46), followed by Phi-3 (4.99) and Mistral-7B (7.81).}
        \label{fig:perplexity_score}
    \end{minipage}
\end{figure}

As shown in Fig.~\ref{fig:performance_metrics}, increasing $L$ from 1 to 3 improves MRR@1 from approximately 0.89 to 0.97 and Precision@3 from approximately 0.33 to 0.37, but the effect is most pronounced on Answer F1-Score, which rises from approximately 0.18 to 0.35 --- nearly a $2\times$ relative gain. This asymmetry indicates that the agentic loop's primary benefit is not marginal re-ranking of already-retrieved passages, but recovering answer-relevant context that single-pass retrieval misses entirely, consistent with our claim that iterative query reformulation is essential for navigating OCR noise and layout ambiguity in this corpus.

To assess how well different instruction-tuned language models comprehend and adapt to the domain-specific historical text and OCR-induced artifacts within our legislative archive corpus, we evaluate the average perplexity scores across three prominent backbones: Mistral-7B, Phi-3, and Qwen2.5 as shown in Fig. ~\ref{fig:perplexity_score}.

\begin{figure}[H]
    \centering
    \includegraphics[width=0.75\linewidth]{images/Ablation_loop.jpg}
    \caption{Iterative retrieval correction through query rewriting. The first two iterations retrieve irrelevant same documents, while the third iteration retrieves the correct document containing the relevant information. The example demonstrates a cross-lingual setting, where the query is in Punjabi and the retrieved document is in English.}
    \label{fig:ablation_iterations}
\end{figure}

\subsection{Limitations}
\label{subsec:limitations}

Despite the strong performance of our multilingual agentic RAG framework, several limitations remain. Severe OCR errors in degraded pages can still affect dense retrieval and entity matching. The iterative agentic loop and LLM-based sufficiency checks also increase inference latency compared with standard retrieval. In addition, complex financial tables and unstructured tabular data may lose hierarchical structure during text-to-chunk conversion. Future work will focus on improving OCR robustness, reducing inference latency, and preserving complex table structures.

\section{Conclusion}
In this work, we introduced an Agentic-RAG framework designed to bridge the gap between complex historical archives and user information needs. By blending iterative query rewriting, adaptive document retrieval, and vision-language generation, our approach effectively tackles the nuances of multilingual document search. 

Evaluated on the Punjab Vidhan Sabha debates dataset—a challenging collection spanning multiple decades in Punjabi, Hindi, and English—the system demonstrated robust resilience against real-world hurdles such as historical vocabulary shifts, degraded text, and severe OCR noise. Unlike traditional, rigid RAG pipelines, the iterative loop implemented here acts as a self-correcting mechanism, dynamically verifying context sufficiency to minimize hallucinations and ground final answers directly in source material. 

Ultimately, this study highlights that agentic workflows can significantly enhance information discovery in multilingual public records. Beyond historical text exploration, the principles demonstrated here pave the way for more dependable, transparent, and accessible digital archives for researchers, policymakers, and the public.

Future work will focus on multi modal retrieval, expanding the system to more Indian languages and government archives, and supporting conversational search with long-context reasoning. User studies with researchers and other archive users will further evaluate its usability and reliability.

\noindent\textbf{Acknowledgment} : This work is supported by MeitY, Government of India, through the NLTM-Bhashini project.We also sincerely thank Prof. Gurpreet Singh Lehal(\url{gslehal@gmail.com}) for his valuable support and assistance during the project evaluation.
\
%
%
%
\bibliographystyle{splncs04}
\bibliography{mybiblography}
\end{document}